\documentclass{article}
\usepackage{spconf,amsmath,graphicx}
\usepackage{amsmath}
\usepackage{amssymb}
\usepackage{multirow}
\usepackage{makecell}

\title{Real-Time Spatio-Temporal Reconstruction of Dynamic Endoscopic Scenes with 4D Gaussian Splatting}
\name{Fengze Li$^{1,\dagger}$, Jishuai He$^{\dagger}$, Jieming Ma$^{1,2,\star}$, Zhijing Wu$^3$ \thanks{$\dagger$ Equally Contributed. \\ }}
\address{\textit{University of Liverpool, Liverpool, UK$^1$},\\ 
\textit{Xi'an Jiaotong-Liverpool University, Suzhou, China$^2$},\\
\textit{University of Cambridge, Cambridge, UK$^3$}}

\begin{document}
%
\maketitle
\begin{abstract}
Dynamic scene reconstruction is essential in robotic minimally invasive surgery, providing crucial spatial information that enhances surgical precision and outcomes. However, existing methods struggle to address the complex, temporally dynamic nature of endoscopic scenes. This paper presents ST-Endo4DGS, a novel framework that models the spatio-temporal volume of dynamic endoscopic scenes using unbiased 4D Gaussian Splatting (4DGS) primitives, parameterized by anisotropic ellipses with flexible 4D rotations. This approach enables precise representation of deformable tissue dynamics, capturing intricate spatial and temporal correlations in real time. Additionally, we extend spherindrical harmonics to represent time-evolving appearance, achieving realistic adaptations to lighting and view changes. A new endoscopic normal alignment constraint (ENAC) further enhances geometric fidelity by aligning rendered normals with depth-derived geometry. Extensive evaluations show that ST-Endo4DGS outperforms existing methods in both visual quality and real-time performance, establishing a new state-of-the-art in dynamic scene reconstruction for endoscopic surgery.
\end{abstract}
\begin{keywords}
3D Reconstruction, Gaussian Splatting, Robotic Surgery, Endoscopic Image
\end{keywords}
\section{Introduction}
\label{sec:intro}
Accurately modeling dynamic scenes from 2D images and rendering lifelike novel views in real-time are critical capabilities in surgical robotics, particularly for endoscopic surgery. Dynamic scene reconstruction significantly enhances the accuracy of robotic-assisted minimally invasive procedures by accurately restoring the spatial and physical properties of human tissues \cite{wang2022neuraendonerf}. This, in turn, contributes to better patient outcomes by supporting higher precision in surgical interventions \cite{vitiello2012emerging, seetohul2023augmented}. 

Advanced techniques such as Neural Radiance Fields (NeRF) \cite{NeRF} and 3D Gaussian Splatting (3DGS) \cite{3dgs} have recently shown remarkable potential in rendering highly realistic static scenes. EndoNeRF was the first to apply NeRF in endoscopic scenes, modeling tissue deformation and canonical density to enhance dynamic scene reconstruction accuracy \cite{wang2022neuraendonerf}. EndoSurf advanced this by introducing signed distance functions to impose self-consistency constraints on neural fields, improving geometric fidelity, though at the cost of extended training times \cite{zha2023endosurf}. SurgicalGS achieves computational efficiency through explicit Gaussian primitives and splatting-based rendering, yet lacking temporal resolution for complex surgical dynamics \cite{chen2024surgicalgs}. However, adapting these methodologies to dynamic endoscopic scene reconstruction presents substantial challenges. 
\begin{figure*}[t]
\centerline{\includegraphics[width=1\textwidth]{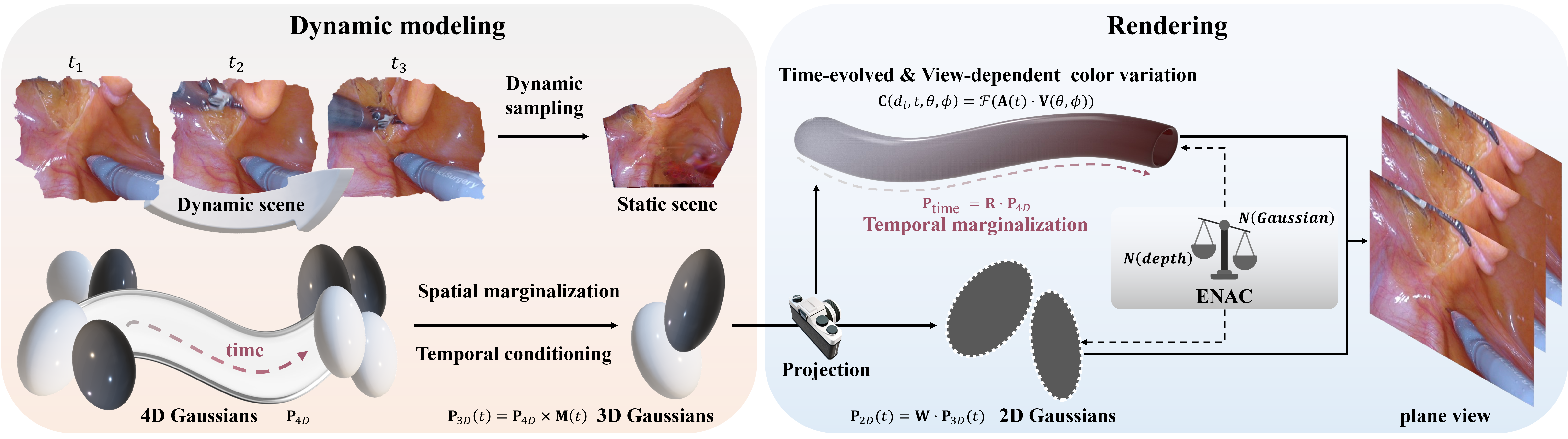}}
\caption{Diagram of the ST-Endo4DGS framework, demonstrating the transition from 4D Gaussians to 2D projections, effectively capturing dynamic endoscopic scenes. Temporal and spatial correlations are preserved through marginalization, with ENAC ensuring alignment of rendered and depth-derived geometry for enhanced fidelity.}
\label{fig01}
\end{figure*}

The continuous motion and temporal dynamics in surgical settings complicate the adaptation of conventional methods to endoscopic applications. As endoscopic procedures capture dynamic scenes through monocular video streams \cite{recasens2021endo}, it is challenging to implement traditional approaches that train static representations per frame and merge them into a dynamic model. Endo-4DGS \cite{huang2024endo} tackled real-time dynamics using a 4D Gaussian deformation field with pseudo-depth data from Depth-Anything \cite{yang2024depth}, enabling real-time reconstruction without ground-truth depth. LGS further optimized 4DGS for surgical settings by incorporating deformation-aware pruning and feature compression, reducing memory and computational costs while maintaining quality \cite{liu2024lgs}. Additionally, EndoGaussian introduced holistic Gaussian initialization to enhance tracking stability and computational efficiency \cite{liu2024endogaussian}. EndoGS integrates depth-guided supervision, spatio-temporal weighting, and surface-aligned regularization to improve geometric fidelity and robustness against occlusions \cite{zhu2024deformable}. Despite these advancements, accurately capturing the spatio-temporal structure of dynamic scenes and managing scale and deformation modeling remain challenging, as many methods still experience local overfitting and limited alignment with true surface geometry.

To address these issues, this study redefines the task by approximating an endoscopic scene’s underlying spatio-temporal 4D volume through a set of 4D Gaussian functions, offering a flexible and unified approach to representing complex endoscopic dynamics \cite{yang2023gs4d}. By enabling 4D rotations, these Gaussians adapt to the 4D manifold, effectively capturing the intrinsic motion in endoscopic scenes. Furthermore, we extend endoscopic spherindrical harmonics, a generalization of 4D spherical harmonics \cite{yang2023gs4d, seeley1966spherical}, to simulate the time-evolved appearance of dynamic endoscopic scenes, allowing for realistic lighting and appearance changes over time. Moreover, inspired by unbiased depth rendering techniques \cite{chen2024pgsr} and normal consistency \cite{fan2024trim}, which are effective in preserving surface detail and reducing local overfitting in complex scenarios, we introduce an endoscopic normal alignment constraint (ENAC) to enhance alignment with true surface geometry, thereby improving the geometric fidelity of dynamic endoscopic scenes. 

Therefore, the primary contributions are: \textbf{(i)} We propose ST-Endo4DGS, a novel framework that utilizes unbiased 4DGS for precise spatio-temporal modeling of dynamic endoscopic scenes, effectively capturing the complex spatial and temporal dynamics of deformable tissues. \textbf{(ii)} We extend endoscopic spherindrical harmonics to model time-evolved appearance, capturing complex lighting and appearance changes in endoscopic scenes with improved temporal fidelity and interpretability. \textbf{(iii)} The endoscopic normal alignment constraint (ENAC) is proposed to enhance alignment between rendered and depth-derived normals, improving the accuracy of monocular depth estimation. \textbf{(iv)} Extensive validation demonstrates that ST-Endo4DGS outperforms existing methods in visual quality and real-time performance.

\section{ST-Endo4DGS}
We introduce ST-Endo4DGS, a framework designed for dynamic endoscopic scene modeling, illustrated in Fig. \ref{fig01}. ST-Endo4DGS utilizes unbiased 4DGS for precise spatio-temporal modeling, extends spherindrical harmonics to capture time-evolving appearance under dynamic lighting, and employs the ENAC constraint to align rendered normals with depth-derived geometry, enhancing depth estimation accuracy. Therefore, this section is organized into four subsections. Subsection \ref{3DGS} reviews 3DGS, establishing its role as a foundational approach for dynamic scene reconstruction. Subsection \ref{ST} presents our 4DGS-based method for comprehensive spatio-temporal modeling. Subsection \ref{ESH} details the extension of spherindrical harmonics for dynamic appearance adaptation, while Subsection \ref{ENAC} discusses ENAC's role in improving geometric fidelity for monocular depth estimation.

\subsection{3DGS as a baseline}\label{3DGS}
3DGS models static 3D scenes using anisotropic Gaussians, enabling real-time, high-fidelity view synthesis through a GPU-optimized rasterizer. This technique has become a foundational baseline for endoscopic scene reconstruction and serves as a key inspiration for our approach. In 3DGS, the scene is represented by a cloud of 3D Gaussians, each affecting a spatial point \( x \in \mathbb{R}^3 \) via an unnormalized Gaussian function:
\begin{equation} 
p(x|\mu, \Sigma) = \exp \left( -\frac{1}{2} (x - \mu)^T \Sigma^{-1} (x - \mu) \right),
\end{equation}
where \( \mu \in \mathbb{R}^3 \) is the mean vector, and \( \Sigma = R S^2 R^T \) is the anisotropic covariance matrix, with \( S = \mathrm{diag}(s_x, s_y, s_z) \) and \( R \) denoting rotation via a quaternion. Each Gaussian includes view-dependent color encoded with spherical harmonics coefficients and an opacity \( \alpha \), both optimized through rendering loss. To refine geometry and quality, 3DGS applies densification and pruning on the Gaussians during optimization.

\subsection{Spatio-temporal modeling with 4DGS}\label{ST}
To capture the complex spatio-temporal dynamics in endoscopic scenes, we extend 3DGS into a 4D framework by integrating the temporal dimension, which enables ST-Endo4DGS to model both spatial structure and temporal evolution, capturing the dynamic nature of tissue deformation and motion. Temporal conditioning is applied to transform a 4D Gaussian into a time-conditioned 3D Gaussian, effectively reducing the 4D representation to a 3D spatial model that varies over time:
\begin{equation}
    P_{3D}(t) = P_{4D} \times M(t),
\end{equation}
where \( P_{3D}(t) \) is the time-conditioned 3D Gaussian, \( P_{4D} \) denotes the complete 4D Gaussian distribution, and \( M(t) \) encodes temporal dependencies. Then, ST-Endo4DGS projects the time-conditioned 3D Gaussian to a 2D Gaussian in the image plane by marginalizing out the depth component:
\begin{equation}
    P_{2D}(t) = W \cdot P_{3D}(t),
\end{equation}
where \( W \) is the projection matrix that maps the 3D Gaussian to the 2D plane, aligning with the view-based rendering process. This step captures the spatial layout while retaining temporal variation. To model time-evolving, view-dependent appearance, we define the color variation \( C(d_i, t, \theta, \phi) \) as a function of both time \( t \) and viewing direction \( d_i \):
\begin{equation}
    C(d_i, t, \theta, \phi) = \mathcal{F} \big( A(t) \cdot V(\theta, \phi) \big),
\end{equation}
where \( A(t) \) is a time-dependent coefficient matrix, \( V(\theta, \phi) \) represents the viewing direction, and \( \mathcal{F} \) captures the interaction between temporal changes and view-dependent lighting effects. This formulation allows ST-Endo4DGS to dynamically adapt appearance based on both time and viewpoint. Temporal marginalization further refines the temporal aspect of the Gaussian model by consolidating the 4D representation into a time-marginalized distribution:
\begin{equation}
    P_{time} = R \cdot P_{4D},
\end{equation}
where \( R \) is a transformation matrix responsible for temporal integration. 

By combining these transformations, ST-Endo4DGS achieves a unified spatio-temporal model that accommodates both static and dynamic properties in endoscopic scenes. The final rendering equation, integrating all components, is expressed as:
\begin{equation}
    \mathcal{I} = \sum_{i=1}^{n} \alpha_i \cdot \big( W \cdot (P_{4D} \times M(t)) \big) \cdot \big( R \cdot P_{4D} \big) \cdot \mathcal{F} \big( A(t) \cdot V(\theta, \phi) \big),
    \label{e6}
\end{equation}
where each term incorporates spatial conditioning, temporal dependency, and view-based color adaptation, enabling high-fidelity spatio-temporal rendering of dynamic endoscopic scenes.

\subsection{Time-evolved appearance with endoscopic spherindrical harmonics}\label{ESH}
In ST-Endo4DGS, capturing the time-evolved and view-dependent appearance of dynamic endoscopic scenes requires a 4D extension of spherical harmonics to incorporate both spatial and temporal dimensions within a unified model. We introduce endoscopic spherindrical harmonics, specifically designed for this purpose, to efficiently encode temporal evolution and angular dependencies within a single representation, minimizing redundancy while ensuring temporal consistency. Thus, we redefine the endoscopic spherindrical harmonics basis function as:
\begin{equation}
Z_{nl}^{m}(t, \theta, \phi) = Y_{l}^{m} (\theta, \phi) \cdot \sin\left( \omega_n t + \phi_n \right),
\end{equation}
where \( Y_{l}^{m}(\theta, \phi) \) captures spatial orientation, and \( \sin(\omega_n t + \phi_n) \) introduces temporal variation. Here, \( \omega_n = \frac{2 \pi n}{T} \) is the angular frequency corresponding to the temporal component with period \( T \), and \( \phi_n \) is a phase shift parameter aligning appearance changes with endoscopic dynamics. This construction establishes an orthonormal basis over the spatio-temporal domain, enabling ST-Endo4DGS to represent time-dependent, view-consistent appearance in a compact and temporally coherent form.

\subsection{Enhancing geometric fidelity with ENAC}\label{ENAC}
The ENAC enhances geometric fidelity by aligning normals derived from rendered depth maps, \( N(\text{depth}) \), with those from the Gaussian representation, \( N(\text{Gaussian}) \). This alignment ensures a precise depiction of surface geometry in dynamic endoscopic scenes. Thus, ENAC introduces an \( L_1 \) regularization term:
\begin{equation}
    L_{\text{ENAC}} = \left| N(\text{depth}) - N(\text{Gaussian}) \right|_1.
\end{equation}

This term acts as a supplementary constraint, refining surface accuracy while supporting the main rendering objective. The rendering model in ST-Endo4DGS, as detailed in Equation~\ref{e6}, integrates both spatial and temporal aspects of the 4D Gaussian representation. The overall optimization objective, balancing rendering fidelity and geometric consistency, is defined as:
\begin{equation}
    L_{\text{total}} = L_{\text{render}} + \lambda_{\text{ENAC}} \cdot L_{\text{ENAC}},
\end{equation}
where \( L_{\text{render}} \) is the main rendering loss, \( L_{\text{ENAC}} \) ensures normal alignment, and \( \lambda_{\text{ENAC}} \) controls its weight in the optimization. Thus, by seamlessly integrating ENAC within ST-Endo4DGS, our approach achieves enhanced alignment with true surface normals, resulting in finer detail and improved coherence across complex endoscopic scenes. 
\begin{figure*}
    \centering
    \includegraphics[width=1\textwidth]{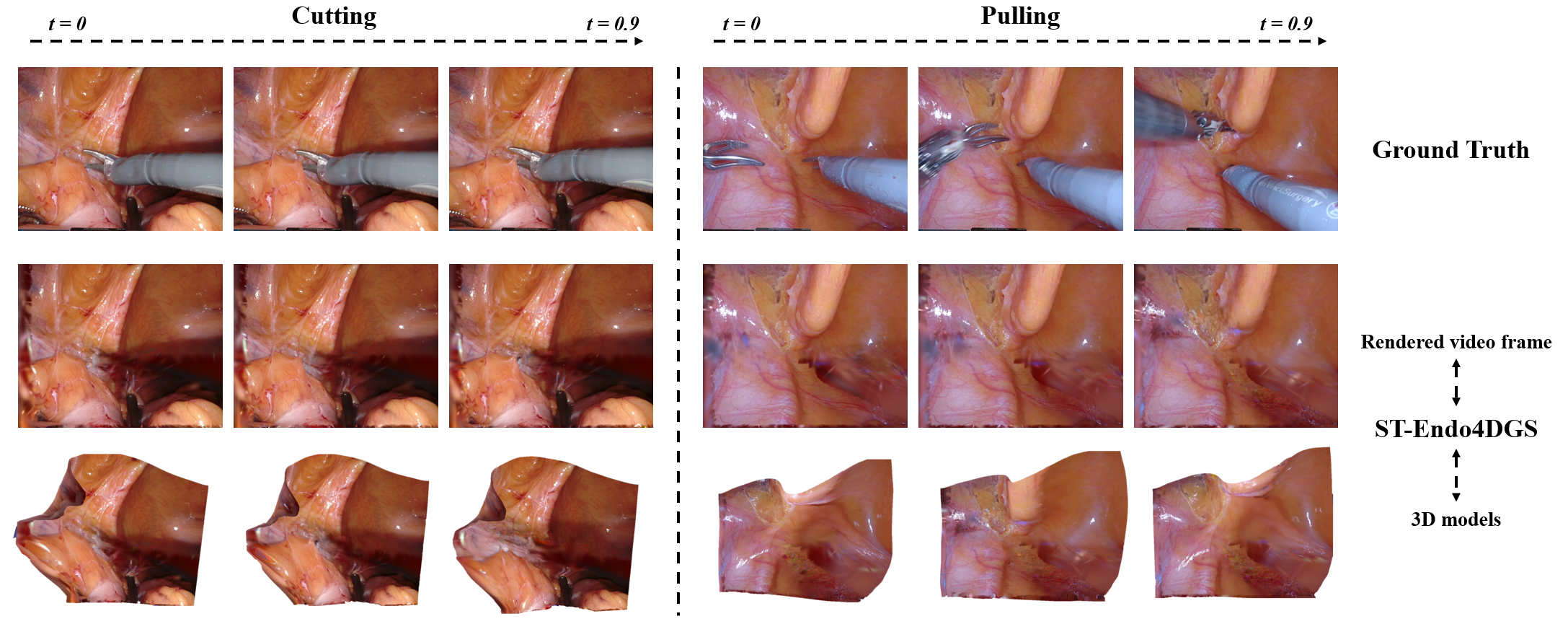}
    \caption{Comparative results on the EndoNeRF dataset showcasing the performance of ST-Endo4DGS under Cutting and Pulling scenarios. The Rendered video frames and 3D models generated by ST-Endo4DGS closely match the ground truth.}
    \label{fig:result1}
\end{figure*}

\begin{table*}[t]
\caption{Comparison and ablation experiments on the EndoNeRF dataset~\cite{wang2022neuraendonerf}, where our models and the best results are in bold.}
\centering
\resizebox{\textwidth}{!}{
\begin{tabular}{c|ccc|ccc|c}
\noalign{\smallskip}\hline
\multirow{2}{*}{Models}
& \multicolumn{3}{c|}{EndoNeRF-Cutting} & \multicolumn{3}{c|}{EndoNeRF-Pulling} & \multirow{2}{*}{FPS $\uparrow$} \\ \cline{2-7}
& PSNR $\uparrow$ & SSIM $\uparrow$ & LPIPS $\downarrow$ & PSNR $\uparrow$ & SSIM $\uparrow$ & LPIPS $\downarrow$ \\ \hline
EndoNeRF~\cite{wang2022neuraendonerf} & 35.84 & 0.942 & 0.057 & 35.43 & 0.939 & 0.064 & 0.2 \\
EndoSurf~\cite{zha2023endosurf} & 34.89 & 0.952 & 0.107 & 34.91 & 0.955 & 0.120 & 0.04 \\
Endo-4DGS \cite{huang2024endo} & 36.00 & 0.950 & 0.039 & 37.71 & 0.960 & 0.041  & 100 \\
EndoGS~\cite{zhu2024deformable} & 37.16 & 0.953 & 0.045 & 36.19 & 0.941 & 0.041 & 70 \\\hline
\textbf{ST-Endo4DGS without ENAC} & 39.17 & 0.950 & 0.039 & 37.78 & 0.960 & 0.041  & \textbf{123} \\
\textbf{ST-Endo4DGS} & \textbf{39.29} & \textbf{0.973} & \textbf{0.016} & \textbf{38.28} & \textbf{0.966} & \textbf{0.024 } & \textbf{123} \\\hline
\end{tabular}}
\label{tab:results}
\end{table*}

\section{Experiments}
We evaluate our model's performance on the EndoNeRF dataset \cite{wang2022neuraendonerf}, which includes two prostatectomy samples captured with stereo cameras, providing depth maps and presenting challenges such as tool occlusions and non-rigid tissue deformations. We follow a 7:1 train-validation split and assess 3D scene reconstruction quality using PSNR, SSIM, and LPIPS metrics. All experiments are conducted on an NVIDIA RTX4090 GPU using PyTorch. Consistent with the Endo-4DGS \cite{huang2024endo} baseline, we adopt a learning rate of $1.6 \times 10^{-3}$.

In our experimental results (shown in Table. \ref{tab:results} and Fig. \ref{fig:result1}), we compare the performance of ST-Endo4DGS against recent models, including EndoNeRF~\cite{wang2022neuraendonerf}, EndoSurf~\cite{zha2023endosurf}, Endo-4DGS~\cite{huang2024endo}, and EndoGS~\cite{zhu2024deformable}, focusing on PSNR, SSIM, LPIPS, and FPS metrics. Our ST-Endo4DGS model demonstrates significant improvements over the counterparts, achieving state-of-the-art performance. Specifically, ST-Endo4DGS achieves up to 6.5\% higher PSNR in the Cutting and Pulling scenarios, with values reaching 39.29 and 38.28, respectively. In terms of perceptual quality, ST-Endo4DGS also records lower LPIPS scores, reflecting a 41.5\% improvement over Endo-4DGS. Furthermore, ST-Endo4DGS operates at 123 FPS, markedly faster than Endo-4DGS’s 100 FPS, representing a 23\% increase in speed. This enhanced efficiency, combined with superior visual fidelity, underscores the advantages of our approach, making ST-Endo4DGS more suitable for real-time applications in dynamic and complex endoscopic environments. Our model not only provides improved image quality but also ensures practical applicability through its computational efficiency.

\section{Conclusion}
We present ST-Endo4DGS, a pioneering framework for dynamic endoscopic scene reconstruction that achieves real-time, high-fidelity novel view synthesis by approximating the spatio-temporal 4D volume through 4D Gaussian primitives. By leveraging unbiased 4D Gaussian Splatting, extended spherindrical harmonics, and the ENAC constraint, ST-Endo4DGS effectively captures intricate tissue deformations and maintains accurate surface alignment. Extensive evaluations on the EndoNeRF dataset confirm the superiority of our approach, with ST-Endo4DGS outperforming existing methods in reconstruction quality, achieving state-of-the-art results. This work establishes a new benchmark for real-time reconstruction of dynamic endoscopic scenes.

\section{Acknowledgments}
This research is supported by the Natural Science Foundation of China (Grant No. 62472361), the Suzhou Science and Technology Project-Key Industrial Technology Innovation (SYG202122), 2024 Suzhou Innovation Consortium Construction Project, the XJTLU Postgraduate Research Scholarship (Grand No. PGRS1906004), the  XJTLU AI University Research Centre, Zooming New Energy-XJTLU Smart Energy Joint Laboratory, Jiangsu Province Engineering Research Centre of Data Science and Cognitive Computation at XJTLU and SIP AI innovation platform (YZCXPT2022103).

\bibliographystyle{IEEEbib}
\bibliography{refs}

\end{document}